# Review of deep learning for photoacoustic imaging

Changchun Yang[+], Hengrong Lan[+], Feng Gao, and Fei Gao[*]

*Abstract—* Machine learning has been developed dramatically and witnessed a lot of applications in various fields over the past few years. This boom originated in 2009, when a new model emerged, that is, the deep artificial neural network, which began to surpass other established mature models on some important benchmarks. Later, it was widely used in academia and industry. Ranging from image analysis to natural language processing, it fully exerted its magic and now become the state-of-the-art machine learning models. Deep neural networks have great potential in medical imaging technology, medical data analysis, medical diagnosis and other healthcare issues, and is promoted in both pre-clinical and even clinical stages. In this review, we performed an overview of some new developments and challenges in the application of machine learning to medical image analysis, with a special focus on deep learning in photoacoustic imaging.

The aim of this review is threefold: (i) introducing deep learning with some important basics, (ii) reviewing recent works that apply deep learning in the entire ecological chain of photoacoustic imaging, from image reconstruction to disease diagnosis, (iii) providing some open source materials and other resources for researchers interested in applying deep learning to photoacoustic imaging.

## I. INTRODUCTION

The advent of the era of big data, the increase in computing power following Moore's Law, and open-source, user-friendly software frameworks [1-4] have made remarkable progress in machine learning, which has aroused great interest in both industry and academia. Data-driven *artificial neural networks*, also known as *deep learning (DL)* [5, 6], are good at discovering complex patterns from massive data to determine the optimal solution in the parameter space. The growing computing power of GPUs allows neural networks to be flexibly expanded from depth [7, 8], width [9], and cardinality [10], forming numerous cornerstone-like models, which have also become the state-of-the-art (SOTA) approaches for a series of problems, exerting its magical power in computer vision, natural language processing and robotics.

Among the many image analysis benchmarks, deep learning rapidly surpasses the traditional machine learning technology, making it reach the most important position in computer vision. At the 2012 ImageNet Large-Scale Visual Recognition Challenge (ILSVRC, [11]), a *convolutional neural network*, AlexNet, easily exceeded the *support vector machine* (SVM) to win the championship with a disparity of more than 10%. That is, since this session, deep learning has replaced SVM, and various innovative neural network models have sprung up. The Top-5 error rate used to measure performance has also reached new lows under fierce competition, gradually surpassing human eye recognition (5~10%), and even close to the Bayes error rate. In addition to shining in computer vision, deep learning also shows its advantages in areas such as natural language processing [12, 13], speech recognition [14, 15][1], and even the field of physics [16-18][2].

By hybrid combination of high contrast of optical imaging and high penetration depth of ultrasound imaging, photoacoustic (PA) imaging, one of the non-invasive medical imaging methods [19, 20], has undoubtedly become an emerging application of deep learning. Therefore, we will review recent research of PA imaging (PAI) with deep learning.

## II. PHOTOACOUSTIC IMAGING

### A. PA fundamental physics

Many literatures have introduced the fundamentals of PAI [21-25], here we only give a brief review of PAI in this section. In PAI, several transducers are used to detect the broadband PA signals, which are excited by a nanosecond pulsed laser light. The initial PA pressure can be expressed as:

$$p_0 = \Gamma_0 \eta_{th} \mu_\alpha F, \qquad (1)$$

where $\Gamma_0$ is the Gruneisen parameter of tissue, $\eta_{th}$ is the energy conversion efficiency from light to heat, $\mu_\alpha$ is the optical absorption coefficient, and $F$ is the local optical fluence.

After the generation of $p_0$, the PA wave propagation in the medium can be described by the following PA equation:

---

[+]Those authors contributed equally to this work.
Changchun Yang, Hengrong Lan, Feng Gao and Fei Gao are with the Hybrid Imaging System Laboratory, Shanghai Engineering Research Center of Intelligent Vision and Imaging, School of Information Science and Technology, ShanghaiTech University, Shanghai 201210, China ([*]corresponding author: gaofei@shanghaitech.edu.cn).
Changchun Yang and Hengrong Lan are also with the Chinese Academy of Sciences, Shanghai Institute of Microsystem and Information Technology, Shanghai 200050, China, and University of Chinese Academy of Sciences, Beijing 100049, China.

[1] DeepMind proposed a deep generative model of raw audio waveforms, and you can try it to mimic human voice: https://deepmind.com/blog/article/wavenet-generative-model-raw-audio.

[2] Modern deep learning methods have also achieved some applications in the field of physics, and these methods are often highly transferable across domains, such as cleaning of interference images in astrophysics [16] and denoising in gravity wave analysis [17], which use a combination of mathematical models and machine learning models to learn the relationship of physics from raw data. These methods can also be used in medical image analysis [18, 19].

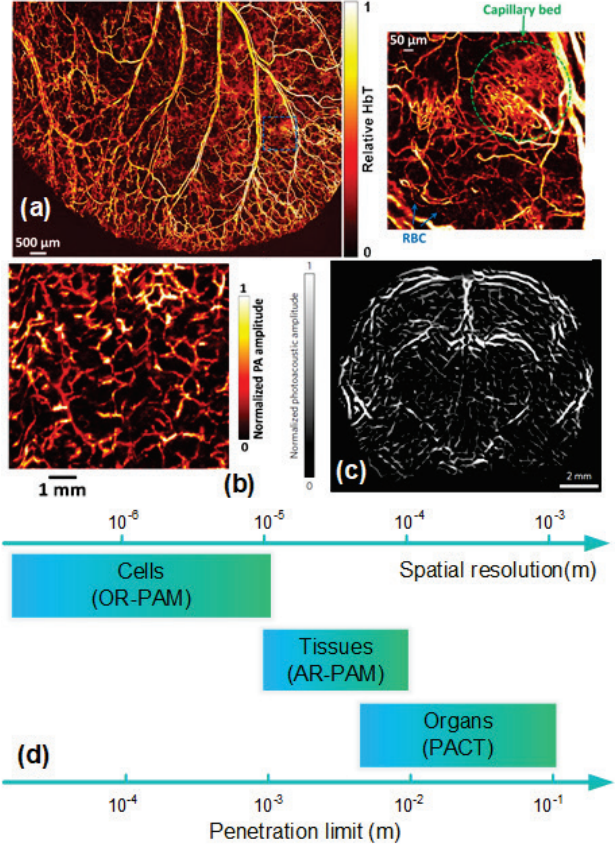

**Fig. 1**. (a) Representative image of OR-PAM imaging a mouse ear, RBC: red blood cell, reproduced with permission from [26]. (b) Representative image of AR-PAM imaging a palm of a volunteer, figures adapted with permission from [27]. (c) Representative image of PACT imaging a mouse head, figures adapted with permission from [31]. (d) The different spatial resolutions and penetration limits for different PAI modalities, figures adapted with permission from [25].

$$\left(\nabla^2 - \frac{1}{v_s^2}\frac{\partial^2}{\partial t^2}\right)p(\mathrm{r},t) = -\frac{\beta}{\kappa v_s^2}\frac{\partial^2 T(\mathrm{r},t)}{\partial t^2}, \quad (2)$$

where $p(\mathrm{r},t)$ is the PA pressure at position r and time $t$, $T$ is the temperature rise, $\kappa$ is the isothermal compressibility, $\beta$ denotes the thermal coefficient of volume expansion, and $v_s$ is the speed of sound. In PAI, the short laser pulse duration should satisfy two confinements: the thermal diffusion time ($\tau_{th}$) and the stress relaxation time ($\tau_s$). Namely, the laser pulse duration should be much less than $\tau_s$ and $\tau_{th}$. The thermal equation can be expressed as:

$$\rho C_V \frac{\partial T(\mathrm{r},t)}{\partial t} = H(\mathrm{r},t), \quad (3)$$

if the laser pulse duration satisfies above condition. The H denotes the heating function, which is defined by the product of the optical absorption coefficient and fluence rate ($H=\mu_a\Phi$). Substituting Eq. (3) into Eq. (2), we obtain the following formula:

$$\left(\nabla^2 - \frac{1}{v_s^2}\frac{\partial^2}{\partial t^2}\right)p(\mathrm{r},t) = -\frac{\beta}{C_P}\frac{\partial H(\mathrm{r},t)}{\partial t}, \quad (4)$$

where $C_P$ denotes the specific heat capacity at constant pressure. We can solve this equation with Green function [21]:

$$p(r,t) = \frac{1}{4\pi v_s^2}\frac{\partial}{\partial t}\left[\frac{1}{v_s t}\int d\mathrm{r}' p_0(\mathrm{r}')\delta(t - \frac{|\mathrm{r}-\mathrm{r}'|}{v_s})\right], \quad (5)$$

where $p_0(\mathrm{r}')$ is the initial pressure at position r′.

### B. PAI modalities

Two fast growing modalities of PAI are photoacoustic microscopy (PAM) and photoacoustic computed tomography (PACT). They obtain images in different ways: the former is point-by-point scanning formed image, and the latter is reconstructed image by acquiring PA signals at different positions.

For PAM, a focused ultrasound transducer is used to scan along the tissue surface. In general PAM setup, both the optical illumination and acoustic detection are confocal. Two typical PAMs are optical-resolution PAM (OR-PAM) and acoustic-resolution PAM (AR-PAM), which depend on either the optical focus or the acoustic focus is finer. OR-PAM can provide a high lateral resolution from a few hundred nanometers to a few micrometers with a high frequency transducer. The PA signal's penetration is limited since the high frequency PA signal suffers severe acoustic attenuation. OR-PAM can provide the vascular anatomy and label-free imaging for hemoglobin oxygen saturation (sO2) as shown in Fig. 1(a) [26]. On the other hand, AR-PAM has a tighter acoustic focus than the optical focus. It achieves resolution of tens of micrometers with acoustic diffraction-limited. The imaging speed of AR-PAM is limited by the scanning speed and the pulse repetition rate of the laser. In Fig. 1(b), AR-PAM can achieve a larger-scale vasculature imaging [27]. Moreover, low scanning speed and small scanning range also limit the applications of PAM.

To accelerate the speed of imaging, PACT uses ultrasonic transducer array to receive PA signals at different positions. An expanded laser beam evenly excites the entire region of interest (ROI), and the PA waves are detected by transducer array simultaneously. After that, reconstruction algorithms, such as universal back-projection (UBP) and time-reversal (TR) [28, 29], are used to reconstruct a high resolution image. PACT with spatial resolution of hundreds of micrometers is preferred, which can be improved by increasing the central frequency and bandwidth of the transducer array. PACT can achieve the whole-body small animals and some organs of human, such as shown in Fig. 1(c) [30, 31].

Generally speaking, the achievable spatial resolution can be estimated if the imaging depth is determined. We list the different spatial resolutions and penetration limits for different PAI modalities in Fig. 1(d).

### III. DEEP LEARNING

It is the core of machine learning to develop some algorithms to make computers capable of learning experience from data to solve tasks. A mathematical model $f$ is created to produce the desired output through training when fed with input data. The training set contains rich data representation information, which can provide experiences for machine learning models. Those models that learned the representation of data are then tuned to produce accurate predictions according to the difference between current predicted output and the expected output by an optimization algorithm. The feedback obtained by training the model on another separate data set, the validation set, is used for further fine-tuning to measure the generalization ability of the model. After iterating through these two steps (tuning and fine-tuning), the model is finally evaluated on the test set to evaluate the performance when it encounters new, unseen data. Tuning, fine-tuning, and evaluation are also the necessary three steps of machine learning to evaluate whether your mathematical model $f$ can solve your task perfectly.

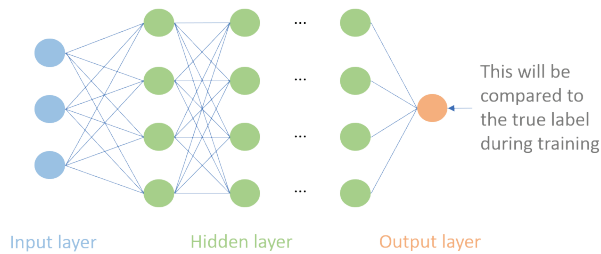

**Fig. 2.** We take the most common artificial neural network, multilayer perceptron or feedforward neural network as an example. The output of the j-th unit at layer i-th is $z_j^i = \theta_j^{iT} x$, where $x$ is the value of the output of the previous layer after the nonlinear function transformation called activation function. Similarly, the output of each layer is passed to the next layer after the activation function, and then the same calculation is performed until the network produces the final output. The training data is fed to the input layer and then the streaming calculation is performed, where the output and derivative of each node are recorded. Finally, the difference between the prediction and the label at the output layer is measured by the loss function. The choice of loss function has a critical impact on the performance of the entire task, so its choice is crucial, common ones include *mean absolute error*, *mean squared error*. You can also manually design the loss function, which is often more attractive. The derivative of the loss function will be used as a feedback signal, and then propagate backward through the network, and the weight of the network will be updated to reduce errors. This is a technique called *backward propagation* [36, 37], which follows the chain rule to calculate the gradient of the objective function with respect to the weights in each node. Then *gradient descent* [38] is used to update all the weights.

We roughly divide machine learning into three categories here according to the training process. First category is the reinforcement learning, the interaction between the constructed agent and the environment achieves the greatest benefit or solves specific problems through learning strategies. One of its most famous applications is AlphaGo [32], which is a Go-playing system developed by DeepMind, even defeating the world's top Go players. The second category is unsupervised learning that is to classify or group based on training samples with unknown categories (unlabeled) and discover patterns between them. A common unsupervised learning example is cluster analysis. The last category is supervised learning, paired data sets make mainstream machine learning algorithms focus on it. Train on the labeled data to find the rules of patterns, and then produce the correct label on the unseen data. For example, many medical imaging including PA image reconstruction problems [33, 34] are based on supervised learning.

Regarding machine learning, especially deep learning, many excellent reviews and surveys have also emerged, you can check the relevant papers [6] for a short introduction, or read free available book [5][3] for an in-depth understanding. See also [35] for a broad understanding of its application to medical image analysis. Only some basic essentials of deep learning will be introduced in this section, serving as useful foundations for its application in PA imaging.

*A. Artificial neural networks*

Firstly we will introduce one of the most famous machine learning models that appeared in the 1950s, the artificial neural network, which is used to briefly explain the deep learning principle. A neural network is composed of many layers, which are connected by many computing units, also known as *neurons*. Data enters the neural network through the input layer, then flows through one or more hidden layers, and finally generates the prediction of the neural network in the output layer, which will be compared with the actual labels (*ground truth*) by an objective function (*loss/cost function*). This difference guides to change the weights of the network to optimize the objective function until the neural network produces a good prediction. This means that the neural network learns experiences from the data and can generalize to new data sets for prediction.

Next, we will briefly explain how the artificial neural networks are constructed. Since the neural network can be very complex and deep, we only introduce its basic form[4] shown in Fig. 2. Map input $x$ to output $y$ through parameterized function $y = f(x;\theta)$, which consists of a lot of nonlinear changes $f(x) = (f_n \circ \cdots \circ f_1)(x)$. Each component/layer $f_k$ can be represented as $f_k = \sigma_k(\theta_k^T f_{k-1})$, followed by a simple linear transformation and a nonlinear function. $\theta_k$ is the model's weights, and the nonlinear functions are differentiable, typically are sigmoid function and ReLUs. The basic idea of training a neural network is simple: training data flows into the network in the forward process, and the gradient of the loss function with respect to every weight is computed using the chain rule, finally *gradient descent* (GD) is used to change these weights to make the loss smaller [36-38]. The reason why such a simple but effective idea did not start explosive growth until recent decade is because that when faced with the exponential growth of parameters and nodes, the calculation will be a huge challenge. Therefore, many techniques[5] such as the design and training of neural networks are open and interesting problems.

---

[3] You can find the free online version: https://www.deeplearningbook.org/.
[4] These are basic but essential concepts and components, compared to *recurrent neural networks* (RNNs), *generate adversarial networks* (GANs), *graph convolutional networks* (GCNs), etc.

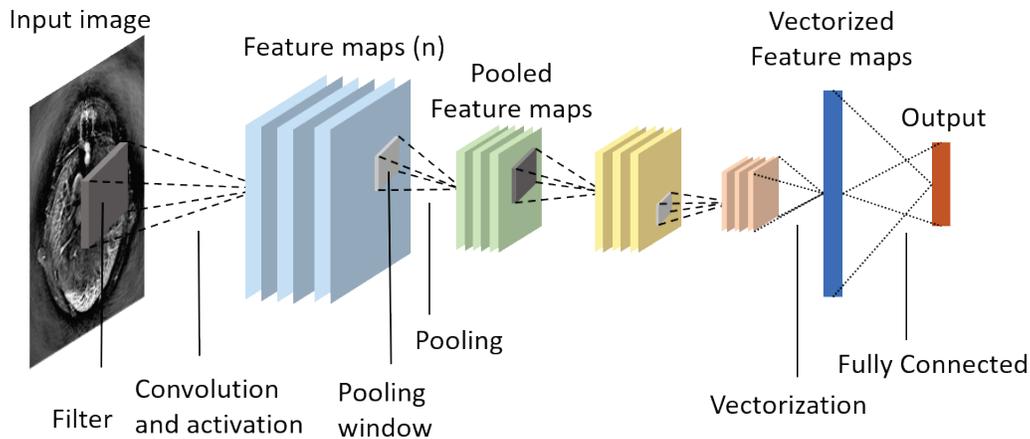

**Fig. 3.** Construction of a typical CNN for medical image analysis, modified according to [49], and input image was taken from [79].

*B. Deep learning*

Traditional machine learning models will extract features from the data manually or with the help of other simple machine learning models before training. However, deep learning will automatically learn representations and features from the data during the training process, without the need for manual design. The main common feature of various deep learning models and their variants is that they all focus on feature learning: automatic learning of data representation, which is also the main difference between modern deep learning and classic machine learning methods. Learning features and performing tasks are improved simultaneously in the model. You can get a comprehensive understanding through the reference [5, 6], or have a detailed study of deep learning with an interactive book [39][6] whose codes use the most popular deep learning framework *pytorch* [3] [7] in academia today.

Deep learning has made rapid progress in medical imaging, mainly relying on *convolutional neural networks* [37] [8] (CNNs). This benefits from the fact that CNNs can easily learn specific features from images or other structured data. Next, we will briefly introduce the essential components of CNNs, understand why it is powerful and have an insight to design your own architecture.

*C. Building CNNs*

Although the introduced artificial neural network can be directly applied to the image, the efficiency of connecting each node of all layers to all nodes of the next layer is very low. Fortunately, structured data such as images can be trimmed and connected based on domain knowledge. CNNs are neural networks that can preserve the spatial relationship between data with few connections. The training process of a CNN is exactly the same as ANNs, except that the structure is often convolutional layers accompanied by activation functions, and pooling layers (as shown in Fig. 3).

(i) **Convolutional layers**. In the convolutional layer, the activations from the previous layer performs a convolution operation with a parameterized filter[9]. Each filter shares weights over the entire scope, which not only can greatly reduce the number of network parameters, but also has the *translational equivariance*. The reason why weight sharing can work is because features that appear in one part of the image may also appear in other parts. For example, this parameterized filter can detect the horizontal line above the image after training to determine the weight, then it can still detect the horizontal line below. After the convolution operation of the convolution layers, a tensor of *feature maps* will be generated.

(ii) **Activation layers.** Similar to the introduction in ANNs, the activation layer is often composed of nonlinear activation functions. Typical activation functions are sigmoid function $\sigma(z) = 1/(1 + e^{-z})$ and rectified linear unit $\text{Re}LU(z) = \max(0, z)$. There are also many other types of activation functions, which can be selected depending on the task and architecture. Because of the existence of these nonlinear activation functions, combined with linear operations such as convolution, the neural network can almost approximate any nonlinear function [40, 41][10]. See [42, 43] to learn more about the role of activation function. The activation layers also generate new tensors of feature maps.

---

[5] Applying deep learning in photoacoustic imaging, recently reported network architecture is significantly more complicated, such as that with multiple outputs [108], the input connected with the output [34], multiple branches [93], multiple loss functions [92], and much more. How to design a novel architecture that can solve specific task is the core.

[6] Free available link: https://d2l.ai/.

[7] Deep learning frameworks undoubtedly facilitate the process of machine learning practitioners turning ideas into code. The well-known ones are: tensorflow [4], pytorch [3], MXNet [2], and caffe [1]. The first two are currently the most commonly used frameworks in industry and academia.

[8] CNN has been used for medical image analysis in the early 1990s, but it was limited by the data and computing power at that time.

[9] It can be compared to the filter in digital image processing, the "convolution" operation here is strictly the correlation operation, but called the "convolution" neural network is also appropriate.

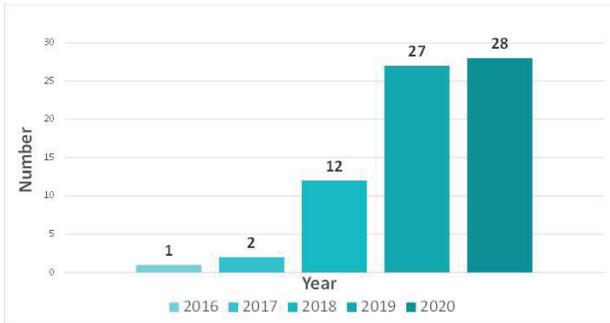

**Fig. 4.** Number of papers of DL in PA imaging in recent years. For 2020 year's data, only the quantity before September 15 is calculated.

(iii) **Pooling**. In neural networks, feature maps are often pooled in the pooling layer. The pooling operation generates a number for each small grid of the input feature maps, which is often achieved by *max-pooling* or *average pooling*. The significance of the pooling layer is to give the neural network *translational invariance*, since a small shift in input will result in changes in the activation maps. For example, to determine whether there is red in the image, translation/rotation will not affect the judgment result. Recent research [44] also proves another alternative to pooling: using convolutional layers with larger stride lengths instead, which can simplify the network structure without reducing performance. *Fully convolutional networks* (FCNs) [45] are also favored by more and more image analysis tasks.

(iv) **Dropout**. A very simple idea has greatly improved the performance of CNNs. Dropout [46] is an averaging technique based on stochastic sampling of neurons to prevent CNNs from overfitting. By randomly removing neurons during the training process, each batch of data uses a slightly changing network, and the weights of the network will be tuned based on the optimization of multiple variations of the network.

(v) **Batch normalization**. Batch normalization (BN) [47][11] is an effective technique to accelerate deep networks training by reducing internal covariate shift. Due to changes in network parameters during training, the distribution of network layer output results is different, making network training difficult. BN can produce normalized feature maps by subtracting the mean and dividing by the standard deviation for each training batch. The data will be periodically changed to zero mean and unit standard deviation with BN, which will greatly speed up the training.

---

[10] No matter how deep the linear operation is, it can still only approximate the linear function without these nonlinear activation functions.

[11] There is no unified conclusion about whether the position of BN is placed after convolution before activation, or after activation. The original paper [47] put it before activation.

In the actual design and improvement of the CNN architecture, above basic components will be combined in a very complex way, with some new and effective operations. When building a CNN for a task, you often have to consider a lot of details to get your CNN perform well on this task. You need to fully understand the task to be solved and find insights to decide how to process the data set before fed to the network. In the early days of deep learning, the construction of modules was often simple. But then more and more complicated structures emerged, and people designed new effective architectures based on previous ideas and insights resulting in updates to the SOTA. See [48] for a survey about recent architectures of CNNs. These novel architectures are often suitable for PA imaging. The structure of most articles we researched can be inspired by them. But before the data is fed into the network, the signal domain[12] or image domain[13] will be processed separately to reduce noise or increase contrast.

**Table 1.** Number of papers of DL in PA based on task.

| Task | Number |
| --- | --- |
| Image reconstruction | 45 |
| Quantitative imaging | 10 |
| Image detection | 7 |
| Image classification | 2 |
| Image segmentation | 6 |
| PA-assisted intervention | 2 |

### IV. Deep Learning For Photoacoustic Imaging

Deep learning has been widely used in clinics to help doctors get a better diagnosis, and successful examples are growing. Because there are too many applications of deep learning in medical imaging of other modalities, we cannot give a comprehensive overview here, but only focus on deep learning in PA imaging. If you are interested, you can see [49-52] for a more thorough review and overview of deep learning in medical imaging, because successful examples often involve multiple organs, and have different ideas and technical details, which is an extremely rich, intersecting, and interesting subject.

Although PA imaging is a new imaging method [22, 23] compared with other modalities of medical imaging, deep learning still shows great application prospects. More specifically, deep learning has been applied at every step of the entire PA imaging workflow. How to obtain high-quality PA image [34] from the sensor data is relevant to the physics of PA. Disease segmentation [53], classification [54], and detection [55] with the reconstructed PA images are relevant to the image domain. Providing functional imaging capability without exogenous contrast: i.e. quantitative imaging of oxygen saturation [56], is also a unique advantage of PA imaging compared to other imaging modalities. PA-assisted intervention is also involved for pre-clinical or even clinical applications [57]. One of the keywords learning/network/Net/learned/convolution/GAN/supervise/un supervise is combined with one of the keywords photoacoustic/optoacoustic to be used for retrieval, then duplicate or unqualified articles (some of them are machine learning) are eliminated. Based on our statistical analysis, the number of papers versus year is shown in Fig. 4. It can be seen that the number on DL used in PA imaging has increased significantly every year. The number classified by category[14] is given in Table 1. Next, we will review in detail from the perspective of each task.

---

[12] Filtering, deconvolution, and interpolation (spatially), etc.

[13] Normalization, standardization, and zero-centered, etc.

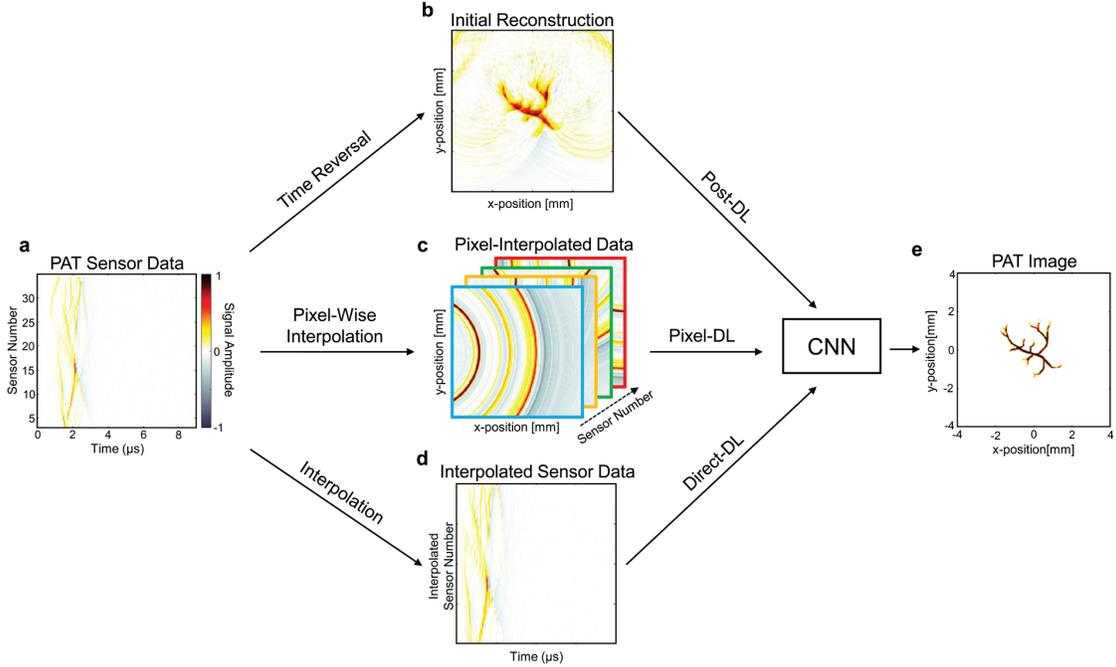

**Fig. 5.** Summary of CNN-based deep learning approaches for PAT image reconstruction. Figures adapted with permission from [65].

*A. PA Image Reconstruction*

Image reconstruction is one of the fundamental components for photoacoustic tomography (PAT), which converts raw signals received by ultrasound transducers to the image of initial pressure distribution. It is a challenging task for PA image reconstruction because of the ill-posed nature and the absence of exact inverse model in practical cases (limited-view and sparse sampling). In PAT, image reconstruction aims to retrieve the initial PA pressure distribution, which indicates the optical absorption of biological tissues. The PA wave propagation at the position **r** and time *t* has the pressure $p(\mathbf{r},t)$ given by Eq. (2). The transducer array receives PA signals excited by a short-pulsed laser light at different locations, which are used to form the PA pressure when *t*=0 by several methods, such as universal back-projection (UBP). We can formulate the UBP reconstruction based on [29]:

---

[14] When counting according to the task category, two papers are about joint task, hence the calculation is repeated twice.

$$p_0(\vec{r}') = \frac{1}{\Omega_0} \int_S d\Omega_0 \times 2 \left[ p(\vec{r},t) - t\frac{\partial p(\vec{r},t)}{\partial t} \right]_{t=\frac{|\vec{r}-\vec{r}'|}{v_s}} \quad (6)$$

where $\Omega_0$ is the solid angle of the entire detection surface *S* with respect to a source point at r. The term of square brackets is the back-projection term. These direct reconstructed algorithms can retrieve satisfactory image if the transducers cover enough angle with sufficient element number to detect the PA signals. However, an ill-posed inverse problem is caused by limited-view or sparse view. Meanwhile, the secondary generated artifacts confuse final PA image.

Model-based reconstruction algorithms show flexibility to be applicable in many unfeasible configurations for direct methods. The inverse problem can be formulated by solving the non-negative least-squares problem:

$$p_0 \in \arg\min \|Ap_0 - y\|^2 + \alpha R(p_0) \quad (7)$$

where $p_0$ is initial PA pressure, *y* is the received PA data from transducers, *A* is the PA forward model, $R(p_0)$ is regularization term, *α* decides the properties between data fit and the regularization term. This problem can be solved by traditional optimization, such as iterative soft-thresholding algorithm (ISTA) [58], GD with repeated iteration. However, these methods are suffering computational complexity and tedious parameters adjustment for different reconstruction tasks. Deep learning methods have found applicability for the reconstruction of PA images, which are driven by a large number of datasets for training. We will review DL-based PA image reconstruction algorithms of two types.

(i) **Non-iterative reconstruction**. Fig. 5 show three different schemes for non-iterative reconstructions: direct estimation,

PA signals and PA image enhancements via deep learning. Direct reconstruction methods only solve the PA wave equation, which take the PA signals as input and capture the mapping from signal to image.

In [59], Dominik Waibel established a direct estimation from raw sensor data for PA imaging. 128-element linear detector's synthetic data were taken into a modified U-net to reconstruct the final initial PA pressure. Similarly, Emran Mohammad Abu Anas [60] proposed a dense CNN architecture for beamforming PA data, which consists of five dense blocks with different dilated convolutions. In this paper, the authors investigated the effect of varying speed of sound for proposed method, and verified the robustness on the speed of sound variation. In PAI, the visibility of deeper object can be affected with the optical scattering, which is a non-negligible issue for the precise depth localization applications. To address this issue, Kerrick Johnstonbaugh [61] designed an encoder-decoder network to predict the location of circular target in deep tissue. This work considered both acoustic and optical attenuation into simulation, and it further closed to the reality. Derek Allman [55] used VGG16 to beamform the raw data to detect the point sources, and removed reflection artifact. A real experiment was also demonstrated in this work. Up to now, all of experiments used synthetic simple (point-like) phantoms. In [62], Johannes Schwab used deep learning to learn the weights in back projection of different channel's data, and the authors used the Shepp-Logan phantom with random enhancement to verify their method. Furthermore, vessel phantom was first used to train neural network in direct reconstruction. He also proposed a data-driven regularization method [63] by applying truncated singular value decomposition (SVD) and then recovering the truncated SVD coefficients, which significantly suppressed noise. Hengrong Lan [64] proposed DU-net, which took three different frequencies' sensor data (2.25MHz, 5MHz and 7.5MHz) as input. In short, DU-net consisted of two U-net, and an additional loss is used to constrain the first U-net. Obviously, the lack of vessel texture structure for raw data, compared with simple phantom, caused a worse result. In [65], Steven Guan also used vessel phantom to compare different reconstruction schemes, and proposed a new type of input in Fig. 5 (from c to e). Jinchao Feng [66] modified Res-Unet to directly reconstruct the simple phantom, and the results are compared with some varietal U-net models. Tong Tong [67] used *in-vivo* data to train the FPnet with a U-net as post-processing. Two types of PA signals (time derivative and normalized original data) are fed into a number of Resblock, and implemented signal-to-image reconstruction by learning below transformation[15]:

$$H_f = F\left(c_2\left(c_0 p(d_i,t) + c_1 \frac{\partial p(d_i,t)}{\partial t}\right)\right) \quad (8)$$

[15] The end of FPnet has a large FC layer, which may cause numerous parameters and overfitting. Some regulations should be applied.

In addition, the authors released their code and data in this study. In summary, these direct estimation models capture priori knowledge that being closely related to conventional procedures or not, as shown in Fig. 5 from a to e. On the other hands, PA signal and image enhancement can be performed in the data domain as preprocessing (from a to d in Fig. 5) or in the image domain as postprocessing (from b to e in Fig. 5) instead of solving the PA wave equation. For this scheme, a signal processing or image processing problem usually exists, which is also important for the imaging result of PAM.

The blurring image may be caused by bandwidth-limited/noise-polluted/spatial-sparse PA signals, even though using a refined reconstruction algorithm. Therefore, PA signal enhancement could obviously improve the PA image quality. Sreedevi Gutta [68] proposed the first neural network to enhance the bandwidth of PA data. A five-layers neural network took one channel bandwidth-limited PA signal as input and predicted full bandwidth signal. Navchetan Awasthi used CNN to de-noise and super resolve (from 50 to 100 detectors) the PA sonogram (sinogram) data in [69], which solve the sparse and low-SNR problem in signal domain. The experiments contain both simulation and *in-vivo* experimental cases. Furthermore, they improve the single channel enhanced scheme in [68] and used U-net to de-noise and enhance the bandwidth limited sonogram data [70]. It resolved two issues, and improved the performance compared with previous works. PA signal preprocessing guarantees the quality of reconstruction by compensating the low SNR and distortion of signal induced by limited bandwidth of transducer in data capture procedure. Spatial-sampling caused issue (limited-view) perhaps should be further explored in image domain.

Stephan Antholzer [71] performed sparsely reconstructed image with residual connected U-net. In [59], the authors also used a U-net to estimate the initial pressure from DAS result images. Likewise, Johannes Schwab [72] used a CNN to reconstruct better PA image by feeding 64 detectors and half-view PA image. Light emitting diode (LED) based PA imaging system suffers from low SNR signal due to limited output power. Therefore, the result of LED-based imaging usually has a worse quality. In [73, 74], Emran Mohammad Abu Anas proposed two different architectures, recurrent neural network based and CNN (Dense block) based approaches, to enhanced the SNR of input low quality PA image. The blood mimicking phantoms' data were collected in these works, and generated ground truth by averaging data. After that, Mithun Kuniyil Ajith Singh [75] used higher laser energy to obtain the ground truth, which are used to train a U-net to enhance the lower SNR image from LED-based imaging system. Ali Hariri [76] proposed MWCNN to enhance the low-fluence LED-based PA image, which replaced the pooling and deconvolution of U-net with discrete wavelet transform and inverse wavelet transform respectively. Ref. [77] also used modified U-net to reconstruct PA image, which divided the model into three parts: feature extraction, artifacts reduction (U-net with residual skipped connection), and reconstruction. Ref. [78] extracted features from limited-view image and trained a VGG-net to match the features between limited-view image and full-view image. Neda Davoudi used U-net to recover the quality of sparse PA image [79], both of simulation and realistic data were trained and tested in this work. Moreover, these data and codes are open access. Seungwan Jeon tried to use U-net based CNN to correct the speed of sound aberration in PA image in [80]. In [81] Steven Guan proposed FD-UNet to remove artifacts caused by sparse data. Meanwhile, Tri Vu [82] designed WGAN-GP to decrease the artifacts in PACT, with both phantom and *in-vivo* results validated. Parastoo Farnia [83] post-processed a TR reconstructed image as input of CNN. Furthermore, deep learning is applied in intravascular PAT, which can enhance the cross-sectional images of the vessel by feeding a TR reconstructed image as input [84]. Guillaume Godefroy [85] used U-Net to resolve the visibility problem due to the limited view and limited bandwidth, and transferred the model from simulation to real experiment. Specially, the ground-truth of experimental data came from the CMOS camera with simple processing. In [86], an in-vivo sheep brain imaging experiment is performed by enhancing the low quality image with U-net. Huijuan Zhang [87] proposed RADL-net to mitigate the artifacts in sparse sampling and limited view problem, which is applied for ring-shaped PACT with simulation and experimental data.

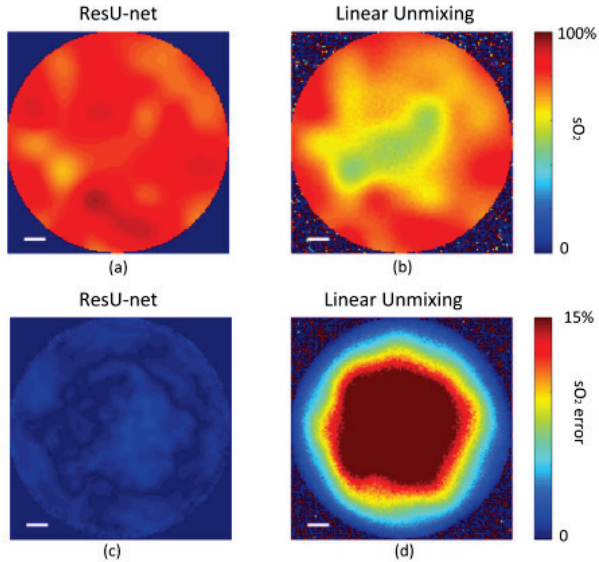

**Fig. 6.** sO$_2$ reconstruction results of deep learning and linear unmixing and the corresponding relative error. Figures adapted with permission from [104].

In PAM, deep learning also plays an important role for image quality correction. In [88], a simple CNN was used to correct motion artifacts in OR-PAM. Israr Ul Haq [89] used a convolutional autoencoder to enhance the quality of image, and the ground truth came from Gobor filter. In [90], Anthony DiSpirito III released a set of mouse brain data, which were used to train a FD U-net to improve the under-sampled PAM images. Simultaneously, Jiasheng Zhou also did a similar work using Squeeze-and-Excitation (SE) block to extract information and verified its effectiveness [91].

In addition, some recent studies open a new scheme neither direct estimation nor pre/postprocessing. Hengrong Lan [92, 93] combined two schemes, and took PA raw data and reconstructed image as input of two novel models. This idea showed better results for limited-view and sparse data. In [65], the authors used intermediate results (pixel-interpolated data) as input of model (from d to e in Fig. 5), where pixel-DL method is proposed. MinWoo Kim pre-delayed every channel's data and got 3D transformed data as input [94]. The improvement of these results is significant using vessel data. The methods that combine the information of different domains and the physical process could be a new frontier of image reconstruction.

(ii) **Iterative reconstruction.** Iterative reconstruction resolves image reconstruction as an inverse problem. In [95], Stephan Antholzer trained a regularization to optimize the compressed sensing PAT reconstruction procedure, and also compared with L$_1$-minimation in [96]. Andreas Hauptmann [34, 97] took a well-known optimization (GD) and corrected the internal optimization by deep learning. The realistic experimental results showed the robustness and superiority of this scheme. Furthermore, Yoeri E. Boink [53] proposed learned primal-dual (L-PD) realizing multi-task (reconstruction and segmentation) in single architecture, which took primal-dual hybrid gradient (PDHG) as a basic optimization. In [19], Changchun Yang used the Recurrent Inference Machines to learn and accelerate the optimization procedure. In other work [98], Hongming Shan simultaneously reconstructed the initial pressure and sound speed distribution, where SR-net was proposed by fusing the gradients of every iteration with previous pressure and sound speed distribution. Navchetan Awasthi proposed PA-Fuse model, which combined two images (BP reconstructed image and model-based reconstructed image) as a fused image [99].

*B. Quantitative imaging*

PAT plays an increasingly important role in both preclinical research and clinical practice. In view of the fact that hemoglobin is one of the major absorbers in human tissue at wavelengths below 1000 nm, PAT can image vascular structure and oxygen saturation of hemoglobin (sO$_2$) by quantifying oxygenated hemoglobin (HbO$_2$) and deoxygenated hemoglobin (HbR). Oxygen saturation is a very important physiological parameter of the human body and can be used to predict the presence of tumors, since the concentration of sO$_2$ of normal tissues is often higher than that of malignant tissues. See [56, 100] for the detailed background of blood oxygenation.

Blood oxygenation, or sO$_2$, is defined as the fraction of HbO$_2$ relative to total hemoglobin concentration in blood:

$$sO_2(x,y) = \frac{C_{HbO_2}(x,y)}{C_{HbO_2}(x,y) + C_{HbR}(x,y)} \times 100\% \quad (9)$$

where $x$ and $y$ is the 2D coordinate of the tissue. The basic reason why we can use PAT for quantitative blood oxygenation imaging is the distinct absorption of $HbO_2$ and $HbR$ at different wavelengths:

$$P(\lambda_i,x,y) = \Phi(\lambda_i)\left(\varepsilon_{HbR}(\lambda_i)C_{HbR}(x,y) + \varepsilon_{HbO2}(\lambda_i)C_{HbO_2}(x,y)\right) \quad (10)$$

where $P(\lambda_i,x,y)$ on the left is the reconstructed 2D PA image at a specific wavelength $\lambda_i$, $\Phi(\lambda_i)$ is wavelength-dependent optical fluence, which is affected by heterogeneous tissue's optical properties and light propagation. $\varepsilon_{HbR}(\lambda_i)$ and $\varepsilon_{HbO2}(\lambda_i)$ are the wavelength-dependent molar extinction coefficients (cm$^{-1}$M$^{-1}$) of HbR and $HbO_2$, which can be obtained from [101] or recorded information [16]. The concentrations of HbR and $HbO_2$ are denoted by $C_{HbR}(x,y)$ and $C_{HbO_2}(x,y)$. Some traditional methods (e.g. linear unmixing) assumes $\Phi(\lambda_i)$ as a constant when solving $sO_2$. In this case, at least two wavelengths are needed to solve the equations, achieving the quantification of $sO_2$. However, $\Phi(\lambda_i)$ is wavelength dependent in real scenarios, so it actually confounds the spectral unmixing of $HbO_2$ and HbR. This leads to the fact that linear unmixing cannot accurately give quantitative results (e.g. shown in Fig. 6 (d)). Solving $sO_2$ corresponding to spatial coordinates from reconstructed PA images of multiple wavelengths is a typical ill-posed inverse problem, for which deep learning is undoubtedly a good choice [102, 103][17].

Chuangjian Cai [104] proposed the first deep learning framework, ResU-net, for quantitative PA imaging. Their model takes the initial pressure images at different wavelengths as inputs, and the outputs are quantitative results of $sO_2$. They use the CNN architecture implemented by U-net to achieve the following functions: (1) determining the contour of the imaged object, (2) performing optical inversion, (3) denoising, (4) determining the type of main absorbing chromophores and estimating their absorption spectrum, (5) unmixing the $HbO_2$ and HbR. Therefore, the author added a residual learning mechanism [7] to solve the gradient explosion problem on the basis of U-net composed of a contraction path that captures comprehensive context and an extension path that realizes precise localization, so that the network can enrich features through deep layers stacking and simultaneously, greatly increase the quantitative accuracy. The author conducted a simulation experiment to simulate a simple numerical phantom based on 21 wavelengths (700-800 nm, step size 5nm) with random gaussian noise. The quantitative comparison results are given in Fig. 6. It can be seen that for linear unmixing, the central region of Fig. 6(d) produced a large estimation error due to spectral coloring caused by light absorption and scattering. In contrast, ResU-net's relative reconstruction error is much smaller (Fig. 6(c)), which implied that deep learning has learned this inverse mapping and can compensate for spectral coloring well. Later, there are some recent works [105-110] that are also based on U-net to quantify $sO_2$. In order to investigate whether it is theoretically possible to give accurate quantitative results using only two wavelengths, Ref. [105] conducted some simulation experiment verifications. They think that the feature maps obtained by deep layer of ResU-net can recursively convolve each row of vectors to obtain sequence information, which is beneficial for blood oxygen saturation of each coordinate in space with contextual information. Clinically-obtained numerical breast phantom is used for optical and acoustic simulations in Ref. [107] avoiding the use of simple layers of tissues. Further, in order to take full advantage of multi-wavelength as inputs, aggregator was added with encoder and decoder. They believe that shallow features are propagated through different stages of aggregation and will be refined [111]. The authors also simply verified the impact on the quantitative results corresponding to the number of wavelengths. When the number of input wavelengths increases, the neural network can produce more accurate results, but the improvement is finally limited. Because the concentration of $sO_2$ in blood vessels is more concerned than other tissues (such as fat), blood vessel segmentation and quantification were performed simultaneously in Ref. [108, 110]. Two symmetric U-nets were used for segmentation and quantification, then focused the analysis result of the entire domain to the segmented blood vessels. Ref. [109] expanded quantitative blood oxygenation from 2D to 3D PA images since the spatial information provided by the 3D images improves the neural network's ability to learn an optical fluence correction and the voxel-wise approach can use the full information. Different from the above work that considers the global spatial information of each wavelength separately, Janek Grohl [112] computed $sO_2$ from pixel-wise initial pressure spectra, which are vectors comprised of the initial pressure at the same spatial location over all recorded wavelengths. Ref. [113] learned the inverse mapping by directly regressing from a set of input spectra to the desired fluence based on eigenspectra multispectral optoacoustic tomography [114]. Both spectral and spatial features were used for inference, which was verified in simulations and experimental dataset obtained from blood phantoms and mice in vivo. This is also the first *in vivo* experiment to prove PA $sO_2$ imaging with deep learning.

The above reviewed methods are all supervised learning for quantitative blood oxygen imaging. However, since the ground truth of $sO_2$ is difficult to obtain for in vivo imaging experiments, it seems that unsupervised learning may be a better choice to solve this problem. Almost all the studies are currently in the stage of feasibility study, Ref. [115] firstly tried to use unsupervised learning to identify regions containing $HbO_2$ and $HbR$. Although these existing works are still at very preliminary stage, we can still see the dawn of quantitative PA imaging using deep learning. With more investigations in clinical use, more realistic data will be obtained and methods will be verified.

---

[16] See https://omlc.org/software/mc/mcxyz/index.html for a detailed tissue properties in the wavelength range of 300~1000 nm.

[17] In fact, deep learning can be used to solve a wide variety of inverse problems arising in computational imaging [102, 103].

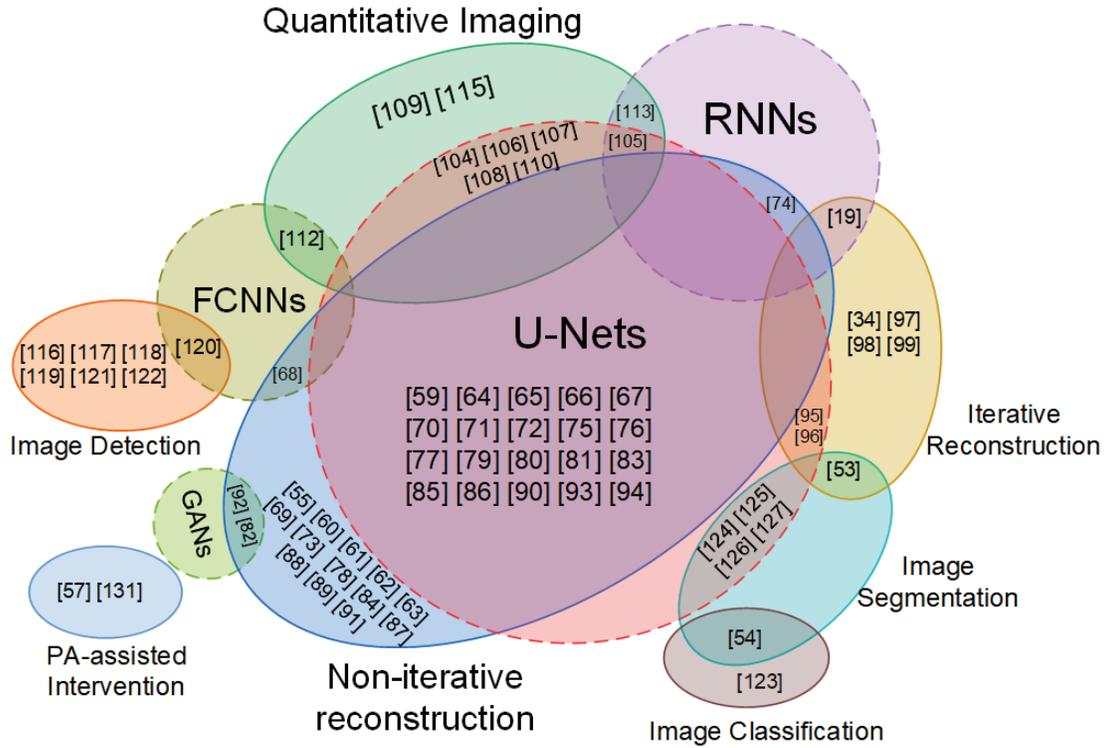

**Fig. 7.** Schematic diagram describing the photoacoustic imaging task and deep learning architecture. The dotted circle represents the network architecture and the solid ellipse represents the task of photoacoustic imaging. References not included in the circle of the four network architectures (U-Nets, GANs, RNNs, FCNNs: fully connected neural networks) in the figure represent the use of other architectures (such as ordinary CNNs).

*C. PA Image detection*

Image detection algorithm is to process the image and detect objects in it. As a promising imaging method, PA imaging, even in the early development stage, also has some published work with deep learning for cancer detection [116-122]. Arjun et. al explored deep neural networks to diagnose prostate cancer based on multispectral PA image database [118]. In this work, a combination of an adaptive greedy forward with backward removal features' selector was used to select features as a key strategy, and then an optimal detection result was given along with a CNN detection/classification model. This is the first time deep learning has been used to detect prostate cancer on *ex-vivo* cancer bearing human prostate tissue with PA imaging. Based on this work, the detection of prostate cancer has expanded to 3D [116], and the experiment was verified based on transfer learning to the prostate using the thyroid database [117]. Ref. [121] used the combined data collected by three wavelengths to learn the discriminative features, and the classifiers are used to judge the benign and malignant prostate cancer, which showed that higher accuracy and sensitivity are higher than the classifier based on the original photoacoustic data. With the help of the multi-wavelength data set, Ref. [122] established and examined a fully automated deep learning framework that learns to detect and locate cancer areas in a given specimen entirely. In order to solve the problem of body hair signal that hampers the visibility of blood vessels in PA imaging, a novel semi-supervised learning (SSL) method was proposed in [119], since the actual training data and test data are very limited. Due to the directional similarity between adjacent body hairs, the author introduced this fact as a priori knowledge into SSL, which enabled the effective learning of the discrimination model for detecting body hair from a small training data set. Finally, the effectiveness of their method was quantitatively verified based on experimental data and successfully applied to virtual hair removal to improve the visibility of blood vessels in PA imaging. Xiang Ma [120] firstly presented the principle of using deep learning to evaluate the average adipocyte size. A deep neural network with fully connected layers was used to fit the relationship between PA spectrum and average adipocyte size. Since the size of adipocyte in obese people is directly related to the risk of metabolic diseases, deep learning has the potential for noninvasive assessment of adipose dysfunction.

Table 2. Open source for deep learning in PA imaging.

| Paper Index | Data |
|---|---|
| [132] | https://anastasiolab.wustl.edu/downloadable-content/oa-breast-database/ |
| [65] | http://www.radiomics.net.cn/post/132 |
| [79] | https://figshare.com/articles/Data/9250784 |
| [89] | https://osf.io/9u3kt/?view_only=36f8fa0e4c884aad80b371b83a993d68 |
| [118] | on request |
| [55] | https://ieee-dataport.org/open-access/photoacoustic-source-detection-and-reflection-artifact-deep-learning-dataset |

| Paper Index | Code |
|---|---|
| [79] | https://github.com/ndavoudi/sparse_artefact_unet |
| [94] | https://drive.google.com/drive/folders/152mhi_f1Toa6tiax4ALs825zGBGGUzcO?usp=sharing |
| [93] | https://github.com/chenyilan/Y-Net |
| [34] | https://github.com/asHauptmann/3DPAT_DGD |
| [55] | https://github.com/derekallman/Photoacoustic-FasterRCNN |

### D. PA Image classification

Image classification is to label objects in the image, firstly detecting the different objects and then classifying them, which can be regarded as an advanced version of image detection. There are only a few preliminary works [54, 123] about classification in PA imaging. Yongping Lin [123] achieved automatic classification of early endometrial cancer based on co-registered PA and ultrasonic (CRPU) signals. The CRPU images were obtained based on Monte Carlo simulations in case standardized CRPU images were hardly got with most PAI systems still in prototype phase. Jiayao Zhang [54] explored the deep learning algorithms for breast cancer diagnostics, where transfer learning was used to achieve better classification performance. Facing the problem of limited data sets, the author used a pre-processing algorithm to enhance the quality and uniformity of input breast cancer images for the experiments.

### E. PA Image segmentation

Image segmentation is the holy grail of quantitative image analysis, the goal of which is outputting a pixel-wise mask of images. The image is divided into several regions or parts based on the characteristics of the pixels. Some recent works have been reported about segmentation of the PA image domain using deep learning [53, 54, 124-127], which is often performed jointly [128-130] [18] with other tasks. A partially-learned algorithm is proposed in Ref. [53] for joint PA reconstruction and segmentation. Different from other reconstruction works, the authors assumed that segmentation of the vascular geometry is of high importance since the visualization of blood vessels is the main task in PAI. The unique advantages of PA imaging make it possible to unmix $HbO_2$ and Hb in multispectral optoacoustic tomography (MSOT), which is beneficial to the precise segmentation. A deep learning approach with a sparse UNET (S-UNET) for automatic vascular segmentation in MSOT images was used in Ref. [124] to avoid the rigorous and time-consuming manual segmentation. The wavelength selection module helped to select the optimal wavelengths for the current segmentation task. It also helped to reduce the longitudinal scanning time and data volume in multi-wavelength experiments. Facing the errors in the co-registration of compounded images of optoacoustic and ultrasound (OPUS) imaging, proper segmentation of different regions turns essential. An automatic segmentation method based on deep learning was proposed in Ref. [125] for segmenting the mouse boundary in a pre-clinal OPUS system. The experimental results were shown to be superior than another SOTA method in a series of experimental OPUS images of the mouse brain, liver and kidney regions. The same authors [126] later proposed an automatic surface segmentation method using deep CNN for whole-body mouse OPUS imaging. This method can achieve accurate segmentation of animal boundaries in both photoacoustic and pulse-echo ultrasound images with good robustness. Yaxin Ma [127] proposed a deep learning framework for automatically generating digital breast phantom. The tissue types are segmented from x-ray and combined with mathematical set operations. Finally, human-like optical and acoustic parameters are assigned to generate the digital phantom for PA breast imaging.

### F. PA-assisted intervention

PA imaging has been developed to guide surgery in some research works [57, 131] benefiting from its real-time imaging capability. Ref. [57] used SOTA deep learning methods to improve image quality by learning from the physics of sound propagation. Those deep learning methods were used in PA-assisted intervention to hold promise for visualization and visual servo of surgical tool tips, and evaluated the distance between critical human structures near the tools used for surgery (for example, serious complications, paralysis or death of the patient will occur when major blood vessels and nerves are injured). Neurosurgery, spinal fusion surgery, hysterectomies, and biopsies, these surgeries and procedures will be beneficially affected clinically. Catheter guidance is often used in cardiac interventions, and PA signal can be generated at catheter tips. It is important to accurately identify the position of the catheter tip because the projection will lack depth information during fluoroscopy. Novel deep learning PA point source detection technique was used [131] to identify catheter tips in the presence of artifacts in raw data, prior to implementing the beamforming steps required to form PA images.

---

[18] Joint task refers to the execution of multiple categories of tasks together, producing information complementarity, or one task may have a supervisory effect on another task. There are many examples [128-130] in medical image analysis.

## V. OPEN SOURCE FOR DEEP LEARNING IN PA IMAGING

Most researchers tend to upload their preliminary manuscript to the arXiv[19] preprint server as soon as possible and share their corresponding code on github[20]. You can also find most public data sets through various repositories, which is also the starting point for many competitions or challenges. Many competitions every year often attract many new participants, who put forward their own methods on the same issue to fight for the best results, which always push the SOTA to a new level. With such a rich variety of open access resources, if finding a problem of interest, everyone can easily start their own research based on the public data set, the method described in the preprint, and the project implemented on github.

There have been many excellent articles summarizing the implementation, data sets and challenges. For example, you can find relevant summaries about a short list of publicly available codes for DL in medical imaging, medical imaging data sets and repositories in [49]. Since we can design or improve our own methods and models inspired by excellent, open-accessed deep learning implementation, we mainly briefly summarize the available data sets in PA imaging for deep learning research and give several examples codes about DL in PA. Although limited by the status quo that PA imaging equipment has not been clinically available, and the number of open data sets is small, we believe that these can still help peers to conduct high-quality research based on these data sets. Existing data sets can be divided into two categories: one is physiologically digital phantom, and the other is experimental PA imaging data. Yang Lou in Ref. [132] proposed a computational methodology using clinical contrast-enhanced magnetic resonance imaging data to generate anatomically 3D realistic numerical breast phantoms, and finally provide three different BIRADS breast density levels' digital breast models named Optical and Acoustic Breast Phantom Database (OA-Breast). In [65], the authors released two real experimental datasets (MSOT-Brain and MSOT-Abdomen) and some numerical simulation data. In [79], the realistic mouse cross-sectional data includes sampling data of different probe numbers can be downloadable. In [89], a set of mouse brain data acquired from OR-PAM are shared. [118] introduces a large sample size of prostate cancer patients gathered using multispectral PA imaging. We take all the papers we reviewed into Fig. 7 for summary, and the index of papers with data and some typical code links is given in Table 2. Through Fig. 7, we can clearly see different network architectures are used for different photoacoustic imaging tasks, and quickly find useful data and codes for your own research through Table 2.

## VI. CHALLENGES AND FUTURE PERSPECTIVES

Deep learning shows its potential for medical imaging in various modalities including PA imaging, and achieve SOTA results when tasked with producing precise decisions based on complicated data sets. But there are still many challenges and limitations that need to be overcome. In addition to its data-driven black box-nature, the methods often used for comparison are based on traditional manual design when introducing deep learning into PA imaging. Nevertheless, neural network-based methods can easily outperform these baselines, which makes this comparison meaningless. Largely due to the inability to obtain open data sets, researchers always verify on their own closed data sets with most of the PA imaging systems still in prototype phase. Therefore, although the research on using deep learning to solve some problems in PAI has improved greatly, there is still a lack of fair comparison on training and testing results using large-scale standardized real data sets. We believe when scientists and clinicians gradually develop the current clinical application standardization for PAI, a large amount of real patient data will be available. These problems will be conquered and the entire PA imaging community will develop more harmoniously. As machine learning researchers and clinicians gain more experience, it will be easier to solve current clinical problems using reasonable solutions. Once there are enough mature solution systems based on mathematics, computer science, physics and engineering entering the daily workflow in the clinic, computational medicine will become mainstream. Machine learning and other computational medicine-based technology ecosystem will eventually be established in most biomedical, and PA imaging will finally find its unique position.

## ACKNOWLEDGMENT

This research was funded by Start-up grant of ShanghaiTech University (F-0203-17-004), Natural Science Foundation of Shanghai (18ZR1425000), and National Natural Science Foundation of China (61805139).

---

[19] Deep learning is moving at a breakneck speed, too fast for the standard peer-review process to keep up. In the past few years, many of the most famous and influential papers in machine learning can only be available as preprints before they were published in conference proceedings. Hence you can see the latest research through the preprint website arXiv: https://arxiv.org/.

[20] Most researchers are often willing to share code and data on github (https://github.com/), which to a certain extent also promotes the rapid development of this community.